\documentclass[runningheads]{llncs}
\usepackage{graphicx}
\usepackage{amssymb}
\usepackage{amsmath}
\usepackage{array}
\usepackage{hyperref}
\usepackage[verbose]{wrapfig}
\usepackage{booktabs} 
\newcommand{\x}{\mathbf{x}}
\newcommand{\y}{\mathbf{y}}
\newcommand{\z}{\mathbf{z}}

\begin{document}
\title{Variational Topic Inference for Chest X-Ray Report Generation}
\titlerunning{Variational Topic Inference for Chest X-Ray Report Generation}
%
\author{Ivona Najdenkoska\inst{1} 
\and Xiantong Zhen\inst{1,2}
\and Marcel Worring\inst{1}
\and Ling Shao\inst{2}}
%
\authorrunning{I. Najdenkoska et al.}
%
\institute{AIM Lab, University of Amsterdam, The Netherlands \email{\{i.najdenkoska,x.zhen,m.worring\}@uva.nl}\and
Inception Institute of Artificial Intelligence, Abu Dhabi, UAE
\email{ling.shao@ieee.org}}
\maketitle              
\begin{abstract}
Automating report generation for medical imaging promises to reduce workload and assist diagnosis in clinical practice. Recent work has shown that deep learning models can successfully caption natural images. However, learning from medical data is challenging due to the diversity and uncertainty inherent in the reports written by different radiologists with discrepant expertise and experience. To tackle these challenges, we propose \textit{variational topic inference} for automatic report generation. Specifically, we introduce a set of topics as latent variables to guide sentence generation by aligning image and language modalities in a latent space. The topics are inferred in a conditional variational inference framework, with each topic governing the generation of a sentence in the report. Further, we adopt a visual attention module that enables the model to attend to different locations in the image and generate more informative descriptions. We conduct extensive experiments on two benchmarks, namely Indiana U. Chest X-rays and MIMIC-CXR. The results demonstrate that our proposed variational topic inference method can generate novel reports rather than mere copies of reports used in training, while still achieving comparable performance to state-of-the-art methods in terms of standard language generation criteria.

\end{abstract}

\section{Introduction}

Chest X-rays are one of the most frequently used imaging modalities in clinical practice. However, interpreting X-ray images and writing reports is laborious and creates an extensive workload for radiologists.
Automated radiology report generation using machine learning techniques has thus arisen to potentially alleviate the burden and expedite clinical workflows. Fundamentally, this can be regarded as translating visual input into textual output, which is broadly known as image captioning \cite{vinyals2015show,xu2015show,lu2017knowing,anderson2018bottom}. It is non-trivial and challenging to transfer this to X-ray reports though, as we need to learn their complex structure and diversity, as well as, to model the uncertainty induced by the varying expertise and experience of radiologists. 

Successful chest X-ray report generation methods mainly follow the neural encoder-decoder architecture \cite{Jing_2018,HRGR_agent,liu2019clinically,Yuan_2019,Xue_Yuan_2018,Xue_Yuan_2019,jing-etal-2019-show,Chen_2020,lovelace-mortazavi-2020-learning}, where a convolutional neural network (CNN) encodes the image into a fixed-size representation and then, sentence by sentence, a recurrent neural network decodes the representation into a report. To enhance this architecture, additional techniques have  been introduced. For instance, \cite{Jing_2018} incorporate a co-attention mechanism to exploit the relationships between visual features and medical tags and uses hierarchical LSTMs \cite{lstm_hochreiter_chmidhuber} to generate multiple sentences. Furthermore, to generate reports with high clinical correctness, \cite{liu2019clinically} proposes to optimize a clinical coherence reward by reinforcement learning. To use the information encoded in both the frontal and lateral views, \cite{Yuan_2019} explores the fusion of multi-view chest X-rays. Another relevant approach exploits the structure of reports by modeling the relationship between findings and impression sections \cite{jing-etal-2019-show}. More recent works \cite{Chen_2020,lovelace-mortazavi-2020-learning} leverage the Transformer \cite{vaswani2017attention} as a more powerful language model to better capture long-term dependencies for sentence generation. 

Despite being the state of the art in terms of benchmark measures, these deterministic encoder-decoder models tend to overfit to the data, producing generic results and making them unable to represent the inherent uncertainty in the reports.
This uncertainty arises from the fact that the reports are written by radiologists with different levels of expertise, experience and expressive styles. 
Naturally, this can yield diversity when several radiologists interpret an X-ray image into a report.
In order to improve their generalizability, it is thus highly important to capture the uncertainty when designing algorithms for report generation.
Probabilistic modeling is able to handle the uncertainty, diversity and complex structure of reports  \cite{kohl2018probabilistic,luo2020analysis} in a well founded way. Instead of simply compressing inputs into fixed-sized deterministic representations, which could cause information loss, adopting stochastic latent variables \cite{kohl2018probabilistic} allows the holistic characteristics of sentences, such as topic, style and high-level patterns, to be explicitly modeled \cite{bowman-etal-2016-generating}, which enables more diverse but controllable text generation \cite{Wang_2019,coscvae20neurips}.

In this paper, we propose variational topic inference (VTI), which addresses report generation for chest X-ray images with a probabilistic latent variable model. In particular, we introduce a set of latent variables, each defined as a topic governing the sentence generation. 
The model is optimized by maximizing an evidence lower bound objective (ELBO) \cite{NIPS2015_8d55a249}. During training, the topics are inferred from visual and language representations, which are aligned by minimizing the Kullback-Leibler (KL) divergence between them. By doing so, at test time the model is able to infer topics from the visual representations to generate the sentences and maintain coherence between them. Finally, we adopt visual attention which enables the model to attend to different local image regions when generating specific words.

Our main contributions can be summarized as follows: (1) We propose a variational topic inference framework to address the radiology report generation problem, which enables diversity and uncertainty to be better handled when generating reports. 
(2) We adopt Transformers to aggregate local visual features with each attention head producing a specific representation for each sentence, which encourages diverse sentences to provide informative and comprehensive descriptions.
(3) We demonstrate that our method achieves comparable performance to the state of the art on two benchmark datasets under a broad range of evaluation criteria.

\section{Methodology}
\subsection{Problem Formulation}
Given the input image $\x$, we aim to generate a report that consists of multiple sentences $\{\mathbf{y}_i\}^N_{i=1}$, which are assumed to be conditionally independent. From a probabilistic perspective, we aim to maximize the conditional log-likelihood:
\begin{equation}
\theta^* = \underset{\theta}{\arg\max} \sum^N_{i=1}  \log p_{\theta}(\y_i|\x), 
\label{eq:1}
\end{equation}
where $\theta$ contains the model parameters and $N$ is the number of sentences in each report. To solve the model, we formulate the report generation as a conditional variational inference problem.

\subsection{Variational Topic Inference}
In order to encourage diversity and coherence between the generated sentences in a report, we introduce a set of latent variables to represent topics $\z$, each of which governs the generation of one sentence $\y$ in the final report (note that the subscript $i$ is omitted for brevity). By incorporating $\z$ into the conditional probability $p_{\theta}(\y|\x)$, we have:
\begin{equation}
\log p_{\theta}(\y|\x) = \int_z \log p_{\theta}(\y|\x,\z)p_{\theta}(\z|\x)d\z,
\label{cond_prob}
\end{equation}
where $p_{\theta}(\z|\x)$ is the conditional prior distribution. 
We define a variational posterior $q_{\phi}(\z)$ to approximate the intractable true posterior $p_{\theta}(\z|\y, \x)$ by minimizing the KL divergence between them: $D_{\rm{KL}}[q_{\phi}(\mathbf{z})||p_{\theta}(\mathbf{z}|\mathbf{x},\mathbf{y})]$.
we arrive at:
\begin{equation}
    D_{\rm{KL}}[q_{\phi}(\mathbf{z})||p_{\theta}(\mathbf{z}|\mathbf{x},\mathbf{y})] = \mathbb{E}[\log q_{\phi}(\mathbf{z}) - \log
    \frac{p_{\theta}(\mathbf{y}|\mathbf{z},\mathbf{x})p_{\theta}(\mathbf{z}|\mathbf{x})}{p_{\theta}(\mathbf{y}|\mathbf{x})}] \geq 0,
    \label{}
\end{equation}
which gives rise to the ELBO of the log-likelihood:
\begin{equation}
    \begin{aligned}
        \log p_{\theta}(\mathbf{y}|\mathbf{x}) &\geq 
         \mathbb{E}[\log p_{\theta}(\mathbf{y}|\mathbf{z},\mathbf{x})]
        - D_{\rm{KL}}[q_{\phi}(\mathbf{z})||p_{\theta}(\mathbf{z}|\mathbf{x})]= \mathcal{L_{\rm{ELBO}}(\theta, \phi)},
        \label{eq:elbo}
    \end{aligned}
\end{equation}
where the variational posterior $q(\z)$ can be designed in various forms to approximate the true posterior.

To leverage the language modality during training, we design the variational posterior as $q_{\phi}(\z|\y)$ conditioned on the ground-truth sentence. 
Based on the ELBO, we derive the objective function w.r.t. a report of $N$ sentences as follows:
\begin{equation}
   \mathcal{L_{\rm{ELBO}}(\theta, \phi)} =  \sum^N_{i=1}\big[\sum^L_{\ell=1}\log p_{\theta}(\y_i|\z^{(\ell)},\mathbf{x}) - \beta D_{\rm{KL}}[q_{\phi}(\z|\y_i)||p_{\theta}(\z|\x)]\big],
   \label{eq:elbo_final_2}
\end{equation}
where $\z^{(\ell)}$ is the $\ell$-th of $L$ Monte Carlo samples, and $\beta$ is a weighting parameter that controls the behavior of the KL divergence. In order to efficiently draw samples and conduct backpropagation, we use the reparametrization trick to draw samples from the variational posterior \cite{kingma2013auto}.
During training, the samples are drawn from the variational posterior distribution $\z^{(l)} \sim q_{\phi}(\z|\y)$, whereas during inference the samples are drawn from the prior distribution $\z^{(l)} \sim p_{\theta}(\z|\x)$. 

\subsection{Implementation using Neural Networks}
\begin{figure}[t]
\includegraphics[width=\textwidth]{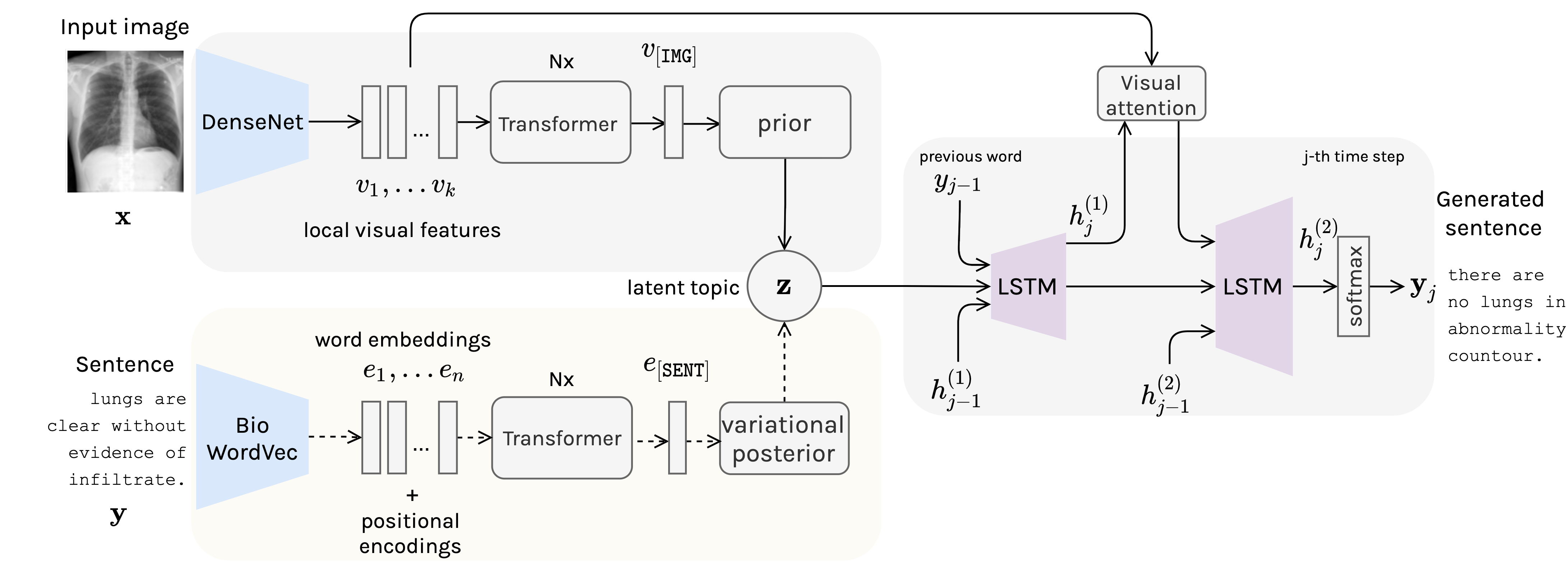}
\caption{Architecture of the proposed variational topic inference model. Note that the language stream is only used at training time, when we infer the distributions of latent topics from both the visual and language modalities and minimize their KL divergence. At test time, we infer topics from the visual modality only to generate sentences.
} 
\label{fig:the_model}
\end{figure}

For efficient optimization, we implement the model with deep neural networks using amortization techniques~\cite{kingma2013auto}. $p_{\theta}(\z|\x)$ and $q_{\phi}(\z|\y)$  are parameterized as fully factorized Gaussian distributions and inferred by multi-layer perceptrons (MLPs), which we refer to as the visual prior net and the language posterior net, respectively. The log-likelihood is implemented as a cross entropy loss based on the output of the sentence generator net and the ground-truth sentence. Figure \ref{fig:the_model} illustrates our proposed VTI model. 

\subsubsection{Visual prior net} To establish more holistic visual representations, we leverage a Transformer to aggregate local visual features from
a pre-trained CNN. Specifically, the convolutional feature maps are flattened along the spatial dimensions to obtain a set of $k$ local visual features $\mathbf{V} = \{\mathbf{v}_1, \mathbf{v}_2, \cdots,\mathbf{v}_{k}\} $, where $\mathbf{v}_i \in \mathbb{R}^{d_v}$ and $d_v$ is the dimension of a visual feature vector. To explore the similarity among local features, we adopt the Transformer to encode them into a special visual token $\mathbf{v}_{[\texttt{IMG]}}$ as the holistic representation of the image. To encourage diversity among topics in a report, we employ a multi-head attention in the Transformer and use each attention head to generate a specific representation for each topic governing the generation of a sentence. 

\subsubsection{Language posterior net}
Each sentence is represented as a sequence of word tokens including a special language token $[\texttt{SENT}]$. Each word token is embedded by an embedding matrix $W_e$, which yields a sequence of $n$ word embeddings \{$\mathbf{e}_1, \mathbf{e}_2, ..., \mathbf{e}_{n}$\}, where $\mathbf{e}_i \in \mathbb{R}^{d_e}$ and $d_e$ is the dimension of the embedding matrix $W_e$.
A Transformer with positional embedding encodes the relationships between the word embeddings, which are aggregated into the special token $\mathbf{e}_{[\texttt{SENT}]}$ as the holistic representation of the sentence. This net takes ground-truth sentences as input to aid the generation of latent topics, which is used during training only.

\subsubsection{Sentence generator net}
The sentences in a report are generated jointly, where the generation of each sentence $\y$ is formulated as $p_{\theta}(\y|\x, \z)$. $\y$ is a sequence of word tokens $\y_0, \y_1, \cdots, \y_t$ and it is common to use the joint probability over the tokens to formulate the generation process: 
\begin{equation}
    p_{\theta}(\y|\x, \z) = \prod^T_{t=1} p_{\theta}\big(\y_t|\x, \z, \y_t\big).
\end{equation}
The sentence generator net is designed in an auto-regressive manner and it is implemented by two consecutive LSTMs \cite{anderson2018bottom} with injected latent topic variables and enhanced by visual attention: 
\begin{equation}
    \mathbf{h}_t^{(1)} = \mathrm{LSTM}^{(1)}\big(\y_t,\mathbf{h}^{(1)}_{t-1},\mathbf{c}^{(1)}_{t-1}\big),
    \label{}
\end{equation}
where $\mathbf{c}^{(1)}$ is the memory cell state initialized by the latent topic $\z$ for the first time step. The next word in the sequence is predicted by the second LSTM, which takes as input the concatenation of the attentive visual representation and the hidden state $\mathbf{h}_t^{(1)}$ of the first LSTM: 
\begin{equation}
    \mathbf{h}^{(2)}_t = \mathrm{LSTM}^{(2)}([\mathbf{v}_{a}; \mathbf{h}_{t}^{(1)}], \mathbf{h}_{t-1}^{(2)}, \mathbf{c}^{(2)}_{t-1}), 
\end{equation}
where $\mathbf{v}_{a}$ is obtained by a visual attention module, as described in the next subsection.
The output  $\mathbf{h}_t^{(2)}$ of the second $\mathrm{LSTM}^{(2)}$ is used to predict the probability distribution $p_t$ of the next word, as in \cite{anderson2018bottom}:
\begin{equation}
    p_t = \mathrm{softmax}(\mathbf{W}_p\mathbf{h}^{(2)}_t),
\end{equation}
where $\mathbf{W}_p \in \mathbb{R}^{d_h \times d_{vocab}}$ is a learnable linear layer that projects $\mathbf{h}^{(2)}_t \in R^{d_h}$ to a probability distribution $p_t$ over the vocabulary of size $d_{vocab}$.

\subsubsection{Visual attention}
To place focus on different parts of the chest X-ray image while decoding the sentence word by word, we use the concept of visual attention \cite{xu2015show}. In particular, the output hidden states $\mathbf{h}_{t}^{(1)}$ of the first LSTM at each time step $t$ are used together with the set of $k$ visual features $\mathbf{V}$ to achieve visual attention. The sum of both representations is fed into a single-layer neural network followed by a softmax function to generate the attention distribution over the $k$ local visual features of the image:
\begin{equation}
    \alpha_{t} = \mathrm{softmax}\big(\mathbf{w}_{a}^\top \mathrm{tanh}(\mathbf{W}_{v} \mathbf{V} + \mathbf{W}_{h} \mathbf{h}_t^{(1)})\big),
\end{equation}
where $\mathbf{w}_{a}^T \in \mathbb{R}^k$, $\mathbf{W}_{v}, \mathbf{W}_{h} \in \mathbb{R}^{k \times d_h}$ are all learnable parameters. Once the attention distribution $\alpha_{t}$ is obtained, we can compute the weighted visual representation as follows:
\begin{equation}
    \mathbf{v}_{a} = \sum_{t=0}^{k} \alpha_{t}\cdot \mathbf{v}_{t},
\end{equation}
which is essentially the aggregated visual representation specific to each word at a given time step $t$.

\section{Experiments}
\subsection{Datasets and Implementation Details} 
We evaluate our VTI model on the Indiana University Chest X-Ray collection \cite{DemnerFushman2016PreparingAC} and MIMIC-CXR \cite{johnson2019mimic} dataset.
Following standard procedure, images are normalized and resized to $224 \times 224$, making them appropriate for extracting visual features from a pre-trained DenseNet-121 \cite{huang2017densenet}. Data entries with missing or incomplete reports are discarded. The impressions and findings sections of the reports are concatenated, lower-cased and tokenized. Non-alphabetical words and words that occur less than a pre-defined threshold are filtered out and replaced with a $[\texttt{UNK}]$ token. Shorter sentences and reports are padded to obtain squared batches. After pre-processing, Indiana U. Chest X-Ray consists of 3,195 samples, which are split into training, validation and test sets with a ratio of 7:1:2. MIMIC-CXR consists of 218,101 samples and is split according to the official splits. 

The word embeddings are initialized with the pre-trained biomedical embeddings BioWordVec \cite{zhang2018}, which represent 200-dimensional contextualized vectors. All hyperparameters are set through cross-validation. The linear layers are initialized from a uniform distribution \cite{he2015delving} and each one has a hidden dimension of 512, followed by ReLU non-linearity and a dropout with a rate of 0.5. The Transformers in both streams use a hidden dimension of 512. The model is trained end-to-end on four NVIDIA GTX 1080Ti GPUs using the Adam optimizer \cite{kingma2017adam} with a learning rate of 3e-05 and early stopping with a patience of five epochs. 
We use cyclical annealing \cite{fu2019cyclical} to deal with the notoriously difficult training with KL divergence in the objective function.
To further improve the readability and coherence, we use a temperature hyperparameter to skew the output distribution towards higher probability events and then apply top-k sampling. 

\begin{table}[h]
\scriptsize
\setlength{\tabcolsep}{2pt}
\begin{center}
\caption{Results on Indiana U. Chest X-ray and MIMIC-CXR using NLG metrics. 
}
\label{tab:NLG_results}
\begin{tabular}{lcccccc}
\multicolumn{7}{c}{\textbf{Indiana U. X-Ray }} \\
\toprule
Method & BLEU-1 & BLEU-2 & BLEU-3 & BLEU-4 & METEOR & ROUGE\\
\midrule
HRGR-Agent \cite{HRGR_agent} & 0.438 & 0.298 & 0.208 & 0.151 & - & 0.322 \\
Clinical-NLG \cite{liu2019clinically} & 0.369 & 0.246 & 0.171 & 0.115 & - & 0.359 \\
MM-Att \cite{Xue_Yuan_2018} & 0.464 & 0.358 & 0.270 & \textbf{0.195} & \textbf{0.274} & 0.366 \\
MvH \cite{Yuan_2019} & 0.478 & 0.334 & 0.277 & 0.191 & 0.265 & 0.318 \\
CMAS-RL \cite{jing-etal-2019-show} & 0.464 & 0.301 & 0.210 & 0.154 & - & 0.362 \\
Memory-Transformer \cite{Chen_2020} & 0.470 & 0.304 & 0.219 & 0.165 & 0.187 & 0.371 \\ 
\textbf{VTI} (Ours)  & \textbf{0.493}  & \textbf{0.360}  & \textbf{0.291}  & 0.154 & 0.218 & \textbf{0.375} \\
\hline
\hline
\multicolumn{7}{c}{\textbf{ MIMIC-CXR }} \\
\midrule
Clinical-NLG \cite{liu2019clinically} & 0.352 & 0.223 & 0.153 & 0.104 & - & 0.307\\
Memory-Transformer \cite{Chen_2020} & 0.353 & 0.218 & 0.145 & 0.103 & 0.142 & 0.277 \\
CC-Transformer \cite{lovelace-mortazavi-2020-learning} & 0.415 & 0.272 & \textbf{0.193} & \textbf{0.146} & 0.159 & \textbf{0.318} \\
\textbf{VTI} (Ours) & \textbf{0.418}  & \textbf{0.293}  & 0.152 & 0.109 & \textbf{0.177} & 0.302 \\
\bottomrule
\end{tabular}
\end{center}
\end{table}

\begin{table}[h]
\scriptsize
\setlength{\tabcolsep}{2pt}
\begin{center}
\caption{Results on MIMIC-CXR using the clinical efficacy metrics.}
\label{tab:CE_results}
\begin{tabular}{lcccccc}
\toprule
\multicolumn{1}{c}{} &
\multicolumn{3}{c}{\textbf{ Micro}} & \multicolumn{3}{c}{\textbf{ Macro }} \\
\midrule
Method & F1 & Precision & Recall & F1 & Precision & Recall\\
\midrule
Clinical-NLG \cite{liu2019clinically} & - & 0.419 & 0.360 & - & 0.225 & 0.209 \\
CC-Transformer \cite{lovelace-mortazavi-2020-learning} & \textbf{0.411} & 0.475 & \textbf{0.361} & \textbf{0.228} & 0.333 & \textbf{0.217} \\
\textbf{VTI} (Ours) & 0.403 & \textbf{0.497} & 0.342 & 0.210 & \textbf{0.350} & 0.151  \\
\bottomrule
\end{tabular}
\end{center}
\end{table}

\begin{figure}[h]
\begin{center}
\includegraphics[width=.95\textwidth]{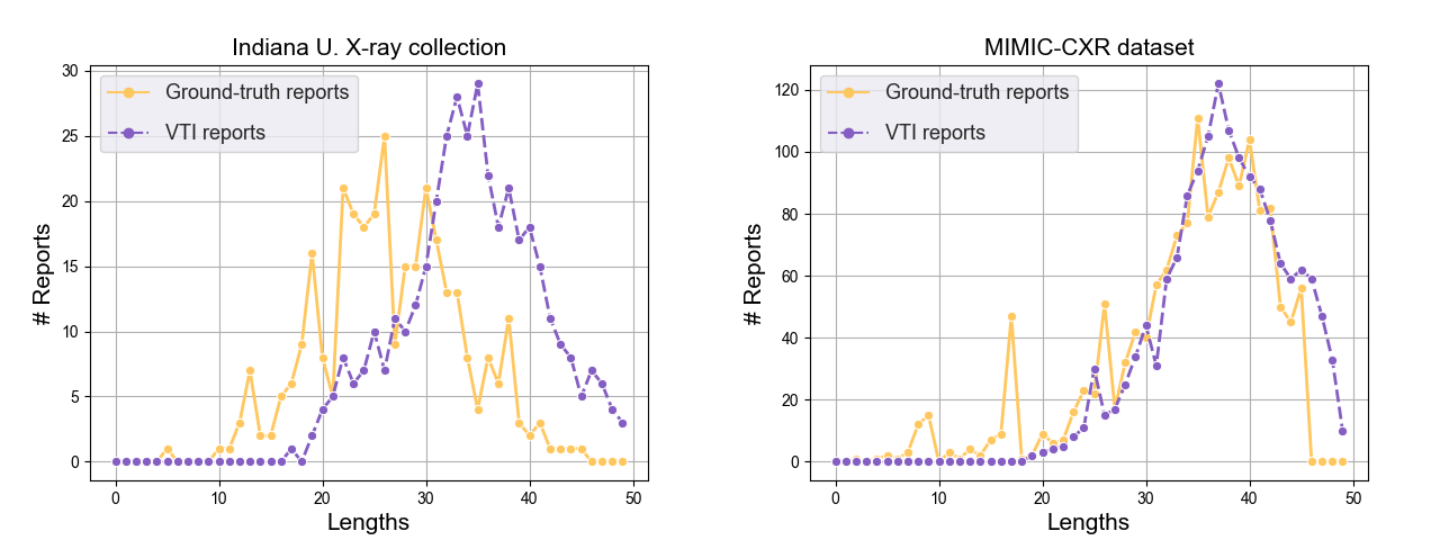}
\caption{Length distributions of the ground-truth reports and the VTI generated reports.} 
\label{fig:lenghts_plot}
\end{center}
\end{figure}

\subsection{Results and Discussion}
We adopt commonly used evaluation metrics for natural language generation (NLG), including BLEU \cite{bleu_paper}, METEOR \cite{meteor_paper} and ROUGE \cite{lin-2004-rouge}. We compare to several other neural network based state-of-the-art methods: \cite{HRGR_agent,liu2019clinically,Xue_Yuan_2018,Yuan_2019,jing-etal-2019-show,Chen_2020} for Indiana U. X-Rays, and \cite{liu2019clinically,Chen_2020,lovelace-mortazavi-2020-learning} for MIMIC-CXR. As shown in Table \ref{tab:NLG_results}, our VTI achieves comparable performance or yields higher scores in terms of BLEU-1-2-3, ROUGE (for Indiana U. Chest X-ray) and METEOR (for MIMIC-CXR). The probabilistic nature of our approach, which imposes diversity, prevents the model from generating longer n-grams similar to the ground-truth, which is important when computing the NLG metrics. 
Our approach is able to maintain a better trade-off between accuracy and diversity, which is desirable when generating descriptions for images, as pointed out in \cite{luo2020analysis}. The benefit of using Transformer encoders to learn holistic representations and BioWordVec for pre-trained word embeddings is empirically observed in the experimental results, details of which are provided in the supplementary material. 

As an additional evaluation in terms of the clinical coherence and correctness, we employ clinical efficacy metrics, i.e., precision, recall and F1 score \cite{liu2019clinically} to compare the extracted labels by the rule-based CheXpert labeler \cite{irvin2019chexpert} for the ground-truth and generated reports. As shown in Table \ref{tab:CE_results}, our model scores higher in precision due to the diversity of the generated reports, which can capture additional information in the image, demonstrating the advantage of probabilistic modeling. 
Moreover, we plot the length distributions of the generated and ground-truth reports, following \cite{Chen_2020}, in Figure \ref{fig:lenghts_plot}. The generated reports tend to be longer for both datasets, suggesting that more detailed information is captured during decoding. They also follow similar distributions, indicating that our VTI is general and not biased towards a particular dataset or simply replicating the exact ground-truth. 

\begin{figure}[t]
\begin{center}
\includegraphics[width=.97\textwidth]{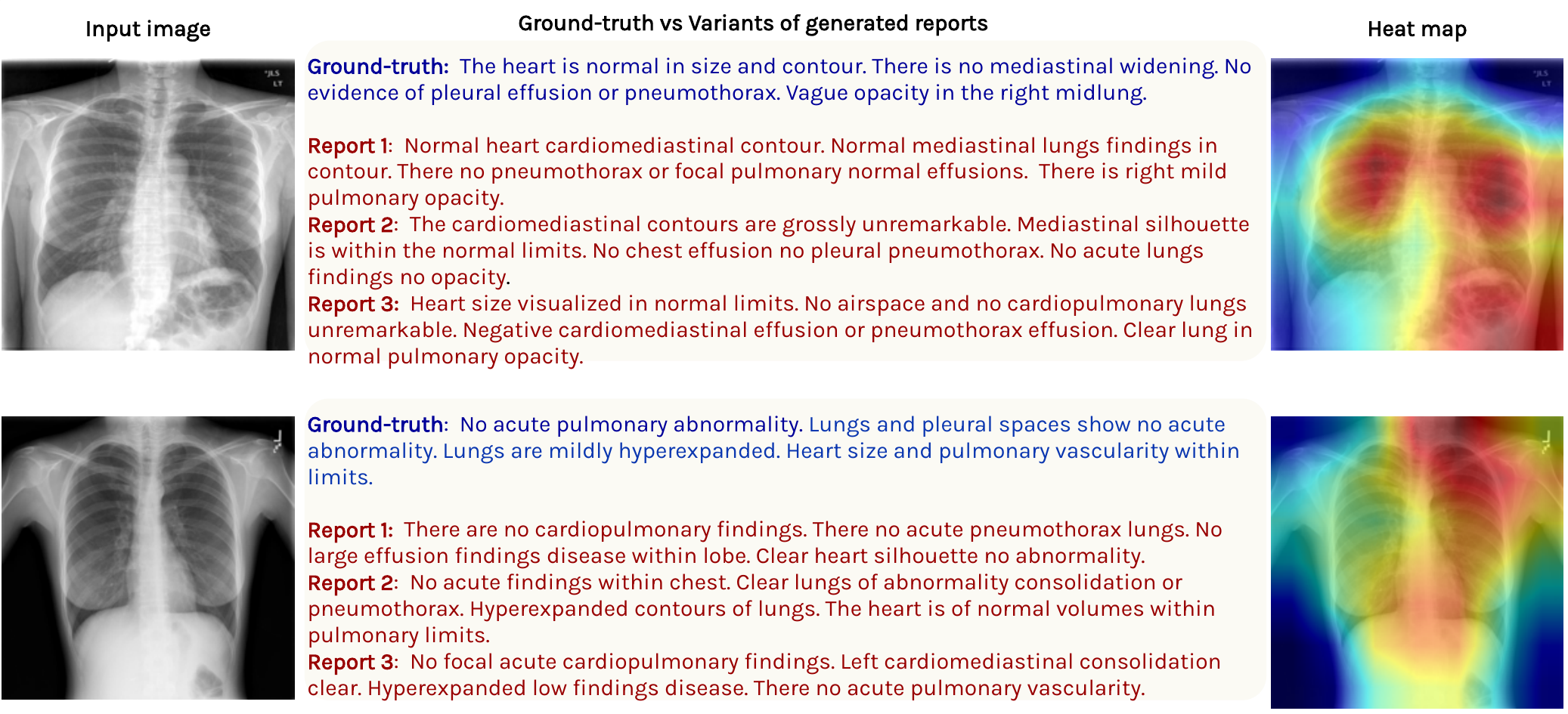}
\caption{Examples of the reports generated for Indiana U. Chest X-rays by our VTI model, and heat maps highlighting relevant image regions using Grad-CAM \cite{selvaraju2017grad}.} 
\label{fig:examples}
\end{center}
\end{figure}

We further examine the results from a qualitative perspective. Specifically, for each image we show three report variants in Figure \ref{fig:examples}, in which we draw one topic sample per sentence, demonstrating that different Monte Carlo samples yield variation in the sentence generation process. We provide heat maps, which show that VTI can focus on relevant image regions while generating the reports. Additionally, visualizations of the attention maps for the generation of each word are presented in the supplementary material.

We notice that the variants describe similar topics with different sentence structures, indicating that the VTI model is aware of more than one correct combination of sentences.
Some sentences have variability in their topics owing to the probabilistic modeling. This tackles the uncertainty in the chest X-ray interpretation process. 
For instance, report 1 in the first image describes the \textit{cardiomediastinal contour} as \textit{normal}, whereas report 2 describes it as \textit{grossly unremarkable}, both with similar semantics. One limitation is that some sentences may have missing words, due to the difficulty of LSTMs in handling long-term dependencies in sequences. This could be alleviated by using a more powerful language decoder, such as Transformer.
Nevertheless, VTI can generate reports not necessarily limited to the ground-truth, indicating its generalization potential, which is a major challenge for report generation \cite{Xue_Yuan_2019,Yuan_2019}. In clinical scenarios, it is often relevant to have a single best report among a variety. The VTI model produces such a report by combining the most probable sentences in terms of Bayesian model averaging in a principled way under the probabilistic framework \cite{kingma2013auto,NIPS2015_8d55a249}.

\section{Conclusion}
In this paper we present a probabilistic latent variable model for automated report generation for chest X-ray images. We formulate the report generation as a variational inference problem. We introduce topics as latent variables to guide the sentence generation by aligning the image and language modalities in a latent space. Our approach allows the diversity and uncertainty that exist in the chest X-ray interpretation process to be modeled. Moreover, it provides a new, theoretically well-grounded framework of probabilistic modeling to deal with common issues in report generation, such as generic, incoherent and repetitive sentences.
We perform extensive experiments on two benchmark datasets, namely Indiana U. Chest X-rays and MIMIC-CXR, and provide a qualitative analysis demonstrating the effectiveness of the proposed model for chest X-ray radiology report generation.

\section*{Acknowledgements}
This work is financially supported by the Inception Institute of 
Artificial Intelligence, the University of Amsterdam and the allowance 
Top consortia for Knowledge and Innovation (TKIs) from the Netherlands 
Ministry of Economic Affairs and Climate Policy.

%
%
\newpage
\bibliographystyle{splncs04}
\bibliography{references}

\end{document}